# Evaluation of Complexity Measures for Deep Learning Generalization in Medical Image Analysis

Aleksandar Vakanski[1] and Min Xian[1]

[1] Department of Computer Science, University of Idaho, Idaho Falls, Idaho, USA
{vakanski, mxian}@uidaho.edu

**Abstract.** The generalization performance of deep learning models for medical image analysis often decreases on images collected with different devices for data acquisition, device settings, or patient population. A better understanding of the generalization capacity on new images is crucial for clinicians' trustworthiness in deep learning. Although significant research efforts have been recently directed toward establishing generalization bounds and complexity measures, still, there is often a significant discrepancy between the predicted and actual generalization performance. As well, related large empirical studies have been primarily based on validation with general-purpose image datasets. This paper presents an empirical study that investigates the correlation between 25 complexity measures and the generalization abilities of supervised deep learning classifiers for breast ultrasound images. The results indicate that PAC-Bayes flatness-based and path norm-based measures produce the most consistent explanation for the combination of models and data. We also investigate the use of multi-task classification and segmentation approach for breast images, and report that such learning approach acts as an implicit regularizer and is conducive toward improved generalization.

## 1 Introduction

The ability of deep neural networks to generalize well, even in cases when overparametrized models are trained on disproportionally small number of data instances, still puzzles researchers [1–4]. Although numerous studies have proposed approaches for explaining this important aspect of deep learning, a solid understanding of the underpinning mechanisms of generalization is still out of reach.

The most common approach has been to derive a *generalization bound* on the excess risk via various learning theories, e.g., PAC-Bayes theory, covering numbers, compression, Rademacher complexity [5–9]. Yet, almost all existing bounds are numerically vacuous in certain settings, and often than not, there is a discrepancy between theoretically predicted bounds and empirical performance. Equally important in understanding generalization are *complexity measures* (*generalization measures*) that correlate monotonically with the generalization performance. Many measures have been proposed either from statistical learning theory bounds and/or empirical studies, e.g., based on the VC dimension [10], norms of model parameters [5, 11], flatness of the loss landscape [12, 13], etc. Although empirical validation of generalization behavior of deep learning models has been reported by a body of works [1, 5, 13–15], further experimental validation is required, by encompassing different datasets, architectural and hyperparameters choices, and optimization algorithms.

Our work was inspired by two recent large empirical studies for evaluating complexity measures. Specifically, Jiang et al. [16] assessed 40 measures using the Network-in-Network (NiN) architecture [17] on CIFAR-10 and Street View House Numbers (SVHN) datasets. The authors in [16] investigated causal relations between various measures and generalization using conditional independence tests, and concluded that many norm-based measures performed poorly. Similarly, Dziugaite et al. [18] used NiN architecture to evaluate 24 complexity measures on CIFAR-10 and SVHN datasets. They argued that empirical validation should be based on the concept of distributional robustness [19], and reported that all complexity measures failed in some experiments. Both studies stated the potential of PAC-Bayes flatness measures.

However, the two studies [16, 18] used large datasets (60K) of non-medical images with 32×32 pixels size, thus, it is not obvious whether their findings will hold for medical image analysis—characterized by small datasets, larger image resolution, and low contrast and signal-to-noise ratio. The dataset size is important, since achieving empirical risk close to 0 may not be the case with medical datasets. Also, the distribution shift in images collected with different devices, protocols, or patient populations poses additional challenges, as it may interfere with the i.i.d. data assumption.

In this paper, we empirically evaluate 25 complexity measures (adopted from [18]) on breast ultrasound images (BUS), using two types of predictor tasks: (i) classification, and (ii) joint classification and segmentation [20–24]. In controlled experiment settings, we vary the depth of the networks to analyze generalization performance. Similar to [16, 18], we observed the potential of flatness- and path-based measures, and differently from these works, we noted a positive correlation with certain spectral norm-based measures. Also, the comparative results showed that the multi-task approach improves generalization on both i.i.d. and o.o.d images, and increases accuracy on small-size datasets. To the best of our knowledge, this is the first study that evaluates a large set of complexity measures for understanding the generalization of single-task and multi-task deep learning networks in medical image analysis.

## 2 Methods

**Notation.** The data distribution over inputs and labels is denoted $\mathcal{D}$. A dataset $\mathcal{S}$ drawn from $\mathcal{D}$ consists of $m$ input-label tuples $\{(\mathbf{X}_1, y_1), \cdots, (\mathbf{X}_m, y_m)\}$, where $\mathbf{X}_i \in \mathcal{X}$ is the input image, and $y_i \in \mathcal{Y} = \{0, 1\}$ is the binary label. A *predictor* (*hypothesis*, *classifier*) $h: \mathcal{X} \to \mathcal{Y}$ maps the inputs to labels. Let $L$ denote the 1-0 classification loss of an input instance. The *risk* (*error*) of the predictor $h$ over the data distribution $\mathcal{D}$ with respect to 1-0 classification loss is $L(h) = \mathbb{E}_{(\mathbf{X},y)\sim\mathcal{D}}[h(\mathbf{X}_i) \neq y_i]$, and the *empirical risk* (*training error*) estimated on the training set $\mathcal{S}$ is $\hat{L}(h) = \frac{1}{m}\sum_{(\mathbf{X},y)\in\mathcal{S}} h(\mathbf{X}_i) \neq$ $y_i$. The difference $\Delta L(h) = L(h) - \hat{L}(h)$ represents the *generalization gap* (*excess risk*) of the predictor $h$. The risk of the predictor $L(h)$ is commonly approximated by the average risk on a test set $\mathcal{S}_{ts}$ consisting of $n$ unobserved input-label tuples drawn from $\mathcal{D}$, i.e., $L(h) \approx \frac{1}{n}\sum_{(\mathbf{X},y)\in\mathcal{S}_{ts}} h(\mathbf{X}_i) \neq y_i$.

**Network architectures.** *Single-task* (*S-T, classification*): The employed classifier has a VGG-like architecture (depicted in blue color in Fig. 1), consisting of multiple blocks (Conv Blocks) with two successive convolutional layers followed by a max-pooling layer. The head portion (FC Block) has 3 fully-connected layers. The generalization performance is evaluated for models with the number of Conv Blocks varying from 2 to 8 blocks. A cross-entropy loss $\mathcal{L}_{\mathrm{CE}} = -\sum y_i log\big(h(\mathrm{X}_i)\big)$ is used for training the network. *Multi-task (M-T, classification + segmentation):* For the combined classification and segmentation task, we added a U-Net-like decoder branch to the encoder part of the classification model (shown in orange color in Fig. 1). The training loss $\lambda\mathcal{L}_{\mathrm{CE}} + (1-\lambda)\mathcal{L}_{\mathrm{DI}}$, is a combination of the cross-entropy loss $\mathcal{L}_{\mathrm{CE}}$ between the predicted and ground-truth classification labels, and the Dice loss $\mathcal{L}_{\mathrm{DI}} = \sum\big(1 - 2|A_{g,i} \cap A_{p,i}|/\big(|A_{g,i}| + |A_{p,i}|\big)\big)$ between the predicted ($A_{p,i} = h_{seg}(\mathrm{X}_i)$) and ground-truth $A_{g,i}$ segmentation masks, weighted by the coefficient $\lambda$.

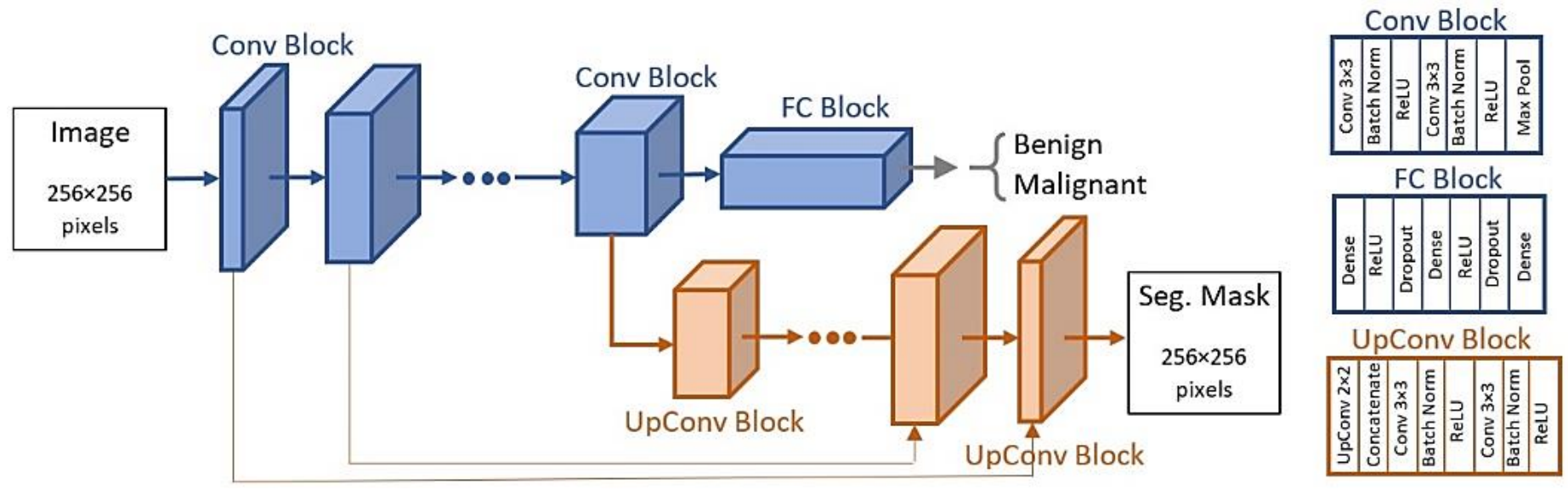

**Fig. 1.** A classification (blue) branch employs a VGG-like architecture with multiple blocks of convolutional and max-pooling layers, followed by fully-connected layers. The segmentation (orange) branch uses a U-Net decoder with upconvolutional blocks and skip connections.

**Experiment design:** Experimental validation of complexity measures was conducted over several experiments $\mathcal{E}$, each encompassing a family of predictors $\{h_1, h_2, \dots, h_k\} \in \mathcal{H}$. The predictors were evaluated by varying the number of Conv Blocks (and the corresponding UpConv Blocks for the M-T approach) in the architecture from 2 (67K parameters) to 8 blocks (152M parameters). Each group of predictors with a fixed architecture (denoted $\mathcal{H}_i$) was run 20 times, typically on different subsets of training and test images (as explained later), using different random seeds for initialization of the trainable parameters. I.e., for each experiment $\mathcal{E}$, the family $\mathcal{H}$ consists of seven groups $\mathcal{H}_{i\in[7]}$ each having 20 predictors $h_{j\in[20]}$, resulting in 140 models. All other hyperparameters were kept fixed, and we used Adam optimization algorithm for all experiments. The motivation behind this choice of experiment design is because the impact of varying the network depth on the generalization is well understood, and it is less ambiguous in comparison to other hyperparameters (such as layers' width, mini-batch size) or the type of optimization algorithm used. We hold that having a prior knowledge of the expected generalization performance is beneficial for this initial study on medical images. Similar to related studies, explicit regularization in the form of weight decay or data augmentation was not used in the experiments, as well as transfer learning via pretraining was not applied.

**Complexity measures.** A complexity measure $\mu$ should capture characteristics of a predictor and/or training data that underpin generalization capacity, i.e., $\mu: (h, \mathcal{S}) \rightarrow \mathbb{R}$. We adopted the complexity measures used by Dziugaite et al. [18]. A brief summary of the measures follows, whereas the mathematical expressions for calculating the measures can be found in the Appendix (full details are provided in [16]). A **VC-dimension based measure** is the VC-dimension of the classifier, in relation to the number of trainable parameters. An **output-based measure** employs the inverse of the margin of the logits values between the binary labels calculated over the training set. **Spectral norm-based measures** are developed using spectral norms of the network parameters [5]. Seven measures are evaluated, which include products, sums, and margin-normalized spectral norms. **Frobenius norm-based measures** are calculated using Frobenius norms of the network parameters, and similar to the spectral norm measures, seven measures are evaluated [16]. **Path-based measures** use squared values of the network parameters, and are calculated as outputs of the predictors on all-ones inputs. We report two related measures. **Flatness-based measures** are based on PAC-Bayes theory and estimate the flatness of the loss landscape in the vicinity of the solution for the network parameters [13, 25]. Six complexity measures are used. An **optimization-based measure** is based on the number of iterations [16] for achieving a training classification error of 0.01 or 0.1, depending on our experiment design (as explained later).

**Correlation between measures and generalization.** *Kendall's sign-ranking coefficient* is frequently used to quantify monotonic variations in a complexity measure $\mu$ and a generalization metric $g$ [16]. The coefficient $\tau$ is calculated as the average sum of pairwise products for all predictors $h_1, h_2$ in a family $\mathcal{H}$ pertaining to $\mathcal{E}$, i.e., $\tau = \left(|\mathcal{H}| \cdot (|\mathcal{H}| - 1)\right)^{-1} \sum_{h_1 \in \mathcal{H}} \sum_{h_2 \in \mathcal{H}} \operatorname{sign}\left(\Delta\mu(h_1, h_2)\right) \cdot \operatorname{sign}\left(\Delta g(h_1, h_2)\right)$. The accepted values for $\tau$ are in $[-1, +1]$ range, where for $\mathcal{E}$ with high positive correlation it is expected that for most predictors in $\mathcal{H}$ if $g(h_1) > g(h_2)$ then $\mu(h_1) > \mu(h_2)$.

*Granulated Kendall's sign-ranking coefficient* was recommended in [16] with the aim of capturing correlations of complexity measures that arise from changing a single hyperparameter between groups of predictors $\mathcal{H}_i$ and $\mathcal{H}_j$. Specifically, the sign-ranking correlation is first calculated between predictors from two different groups $\mathcal{H}_i$ and $\mathcal{H}_j$, $\psi_{i,j} = \left(|\mathcal{H}_i| \cdot |\mathcal{H}_j|\right)^{-1} \sum_{h_1 \in \mathcal{H}_i} \sum_{h_2 \in H_j} \operatorname{sign}\left(\Delta\mu(h_1, h_2)\right) \cdot \operatorname{sign}\left(\Delta g(h_1, h_2)\right)$, and afterward, a granulated coefficient for a measure $\mu$ is calculated by averaging the values $\psi_{i,j}$ over all pairs of predictor groups in $\mathcal{H}$, i.e., $\Psi = (|\mathcal{H}|)^{-1} \sum_{(\mathcal{H}_i, \mathcal{H}_j) \in \mathcal{H}} \psi_{i,j}$. As explained earlier, in our work, one group of predictors corresponds to models with the same architecture (i.e., Conv Blocks).

*Robust Kendall's sign-ranking coefficient* was proposed in [18], based on distributional robustness. For a sign-error $\varphi_{i,j} = (1/2) \sum_{h_1 \in \mathcal{H}_i} \sum_{h_2 \in H_j} \kappa(h_1, h_2) \Big(1 -$ $\mathrm{sign}\big(\Delta\mu(h_1,h_2)\big)\cdot\mathrm{sign}\big(\Delta g(h_1,h_2)\big)\Big)$, the robust coefficient applies larger weights $\kappa$ on samples with small excess risk, based on $\kappa(h_1,h_2)=\max(0,\chi(\Delta g,n)-t)$. The Hoeffding inequality is employed to bound the precision of the samples via $\chi(\Delta g,n)=\big(1-2\cdot exp(-2n(\Delta g/2)^2)\big)^2$. Samples with $\chi(\Delta g,n)$ greater than a threshold $t$, as well as groups $\mathcal{H}_i$ with few remaining samples, are discarded. Such a weighted metric improves the robustness in comparison to the granulated coefficient.

# 3 Experimental Results

**Datasets.** We used two datasets of breast ultrasound images, respectively referred to as BUSIS and BUS-Combined. The **BUSIS dataset** [26] consists of 562 images, of which 306 images contain benign and 256 contain malignant tumors. The **BUS-Combined** dataset includes 3,574 images, created by combining four smaller datasets: BUSIS (562 images), BUSI (647 images) [27], B (163 images) [28], and HMSS (2,202 images) [29]. The dataset has 1,842 benign and 1,732 malignant images. The combination of images from the same BUS domain collected under different settings and/or populations may be considered a violation of the i.i.d. assumption.

**Experiment 1** ($\mathcal{E}_1$): A family of S-T classifiers $\mathcal{H}$ was trained and evaluated on the BUS-Combined dataset (3,574 images). The same subsets of train (80%) and test (20%) images were used for all predictors. Adam optimizer with a constant learning rate of $10^{-5}$ and a mini-batch of 2 images were used for all predictors. The networks were trained until achieving empirical risk of 0.01, similar to the recommendation in [16] (where a training cross-entropy loss of 0.01 was used as a convergence criterion). The networks in this experiment do not use regularization via Batch Normalization or Dropout layers. The means and standard deviations over the predictors' train and test errors are shown in Fig. 2a. We can observe improved generalization as the number of parameters was increased, implying that the optimizer applied implicit regularization. Kendall's coefficients and granulated Kendall's coefficients between the complexity measures $\mu(\mathcal{H})$ and corresponding excess risk $\big(g(\mathcal{H})=\Delta L(\mathcal{H})\big)$ for the family of predictors $\mathcal{H}$ are presented in the first and second columns in Table 1. The results of robust sign-ranking analysis are provided in the Appendix, due to space limitation.

**Experiment 2** ($\mathcal{E}_2$): A family of S-T classifiers $\mathcal{H}$ was trained on the BUSIS dataset (562 images). The data was split into five folds of train (60%), test (20%), and validation (20%) images. Each predictor with the same network architecture $\mathcal{H}_i$ was run 4 times on each fold, resulting in 20 models per group. The learning rate was set to $10^{-5}$. Different from $\mathcal{E}_1$, $\mathcal{E}_2$ was designed to yield models as trained in current practice. I.e., explicit regularization with early stopping was applied: the adopted stopping criterion was non-decreasing risk on the validation subset for 30 epochs. Also, additional regularization was enforced through Batch Normalization and Dropout layers in the network architectures. The train and test errors are shown in Fig. 2b (see also Fig. 3a, and Table A3 in Appendix). Kendall's correlations of the complexity measures with the excess risk $\Delta L(\mathcal{H})$ are displayed in the third column in Table 1. For this type of experiment design, one can notice that the excess risk $\Delta L(\mathcal{H})$ in Fig. 2b increases as the depth of the networks increases. This is due to the underfitting of the predictors with 2 or 3 Conv blocks, which results in a small generalization gap for such predictors. Also, the train error does not reduce to almost 0, as in $\mathcal{E}_1$. Such outcomes do not support the widely reported findings that overparametrized deep learning models improve generalization: i.e., it is expected that the VC dimension (num.params in Table 1) is negatively correlated with the generalization gap $\Delta L(\mathcal{H})$. Therefore, for this experiment we also calculated Kendall's correlation of the complexity measures with the risk on the held-out test subset (that is, for $g(\mathcal{H}) = L(\mathcal{H})$), which is shown in the fourth column in Table 1. The obtained ranking coefficients for this case are more consistent with the reported observations in the literature, since the prediction error $L(\mathcal{H})$ reduces as the depth of the networks is increased.

**Experiment 3** ($\mathcal{E}_3$): This experiment employs M-T learning approach on the BUSIS dataset (562 images). The aim of $\mathcal{E}_3$ is improved classification, by leveraging the segmentation branch for refining and speeding up the feature representation learning. The classification branch of the networks is identical to $\mathcal{E}_2$. The Dice loss for the segmentation branch $\mathcal{L}_{\mathrm{DI}}$ obtains larger values than the classification cross-entropy loss $\mathcal{L}_{\mathrm{CE}}$, therefore, we set the weighting coefficient to $\lambda = 0.2$. The learning rate for $\mathcal{E}_3$ was set to $10^{-4}$, due to the different training loss used in comparison to $\mathcal{E}_2$. The stopping criterion was non-decreasing classification risk on the validation subset for 30 epochs. The resulting risks are shown in Fig. 2c (see also Fig. 3a), where for consistency with $\mathcal{E}_1$ and $\mathcal{E}_2$, the number of parameters in Fig. 2c are shown only for the classification branch. For a trained M-T model, complexity measures were calculated only for the classification branch. The obtained correlation coefficients for the excess risk $\Delta L(\mathcal{H})$ and test set risk $L(\mathcal{H})$ are shown in columns 5 and 6 in Table 1. The optimization-based measure for $\mathcal{E}_2$ and $\mathcal{E}_3$ was recorded for attaining a training error of 0.1 (the predictors that did not achieve 0.1 error were ignored for this measure).

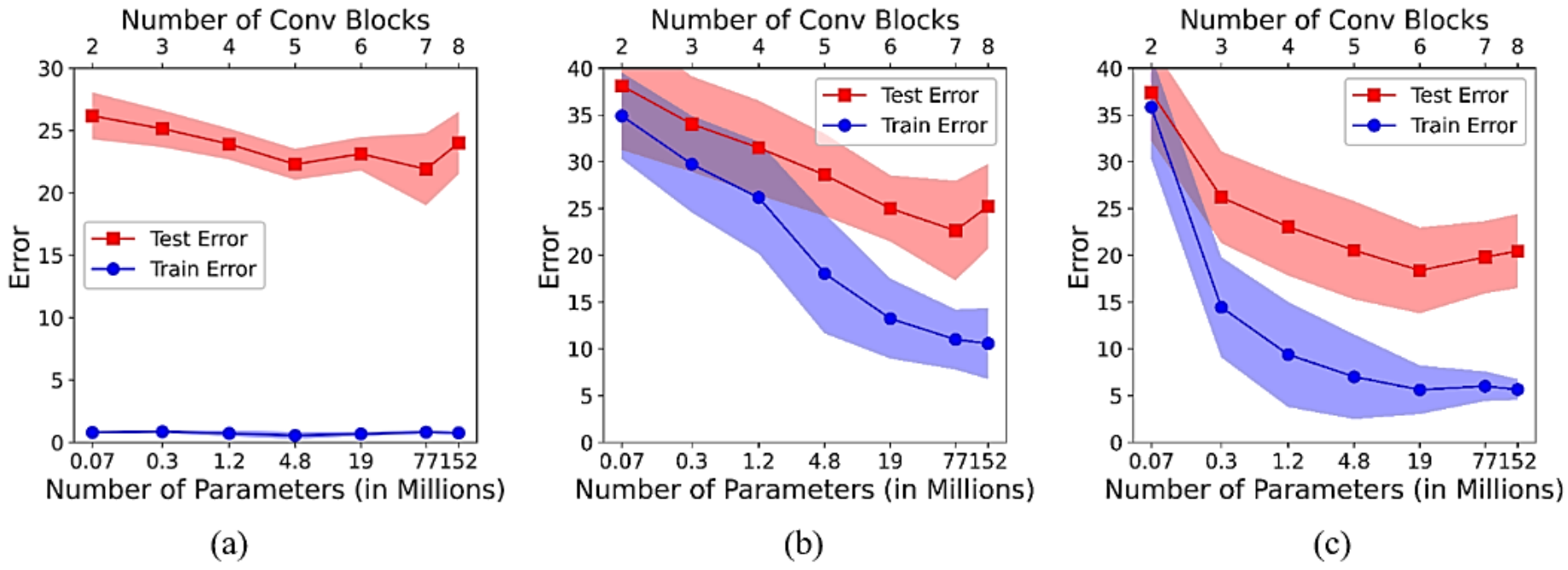

**Fig. 2.** Train and test errors for: (a) S-T classifier $\mathcal{E}_1$ without regularization trained until 1% train error. (b) S-T classifier $\mathcal{E}_2$ trained until non-decreasing validation error. (c) M-T predictor $\mathcal{E}_3$ trained until non-decreasing validation error.

**Imbalanced data metrics.** It is common for medical image datasets to have class imbalance (e.g., a smaller portion of all examined tumors are malignant). Consequently, the average 1-0 accuracy may be misleading in some cases, and other metrics are often used. Fig. 3b displays the F1 score, specificity, and sensitivity for the test subset of BUSIS images. The results reveal that M-T learning improved the performance over these three metrics.

**Cross-distribution evaluation.** We further evaluated the generalization capacity of the predictors on different data distributions from the same domain. In particular, S-T and M-T predictors trained on the BUSIS dataset were evaluated on BUSI, B, and HMSS datasets. The classification accuracies are depicted in Fig. 3c (see Table A4 in Appendix). As expected, the performance decreased significantly on o.o.d. images. For all datasets, M-T approach achieved improved generalization and reduced risk, and required lower number of parameters for achieving comparable results to S-T.

**Table 1.** Kendall's rank correlation coefficients between complexity measures and generalization performance. Color code: low 0–0.3; moderate 0.3–0.5, and high correlation 0.5–1.

| Complexity Measure | Exp 1 $\Delta L$ | Exp 1 $\Delta L$ $\Psi$ | Exp 2 $\Delta L$ | Exp 2 $L$ | Exp 3 $\Delta L$ | Exp 3 $L$ |
|---|---|---|---|---|---|---|
| **VC dimension** | | | | | | |
| num.params | -0.280 | -0.299 | 0.331 | -0.485 | 0.430 | -0.394 |
| **Output measure** | | | | | | |
| inverse.margin | 0.290 | 0.325 | -0.320 | 0.281 | -0.375 | 0.349 |
| **Spectral-norm measures** | | | | | | |
| log.prod.of.spec | 0.406 | 0.416 | 0.275 | -0.516 | 0.534 | -0.394 |
| log.prod.of.spec.over.margin | 0.457 | 0.480 | 0.206 | -0.586 | 0.169 | -0.133 |
| log.sum.of.spec | 0.448 | 0.464 | -0.435 | 0.305 | 0.303 | -0.189 |
| log.sum.of.spec.over.margin | 0.462 | 0.485 | -0.412 | 0.305 | -0.262 | 0.255 |
| log.spec.init.main | 0.462 | 0.480 | 0.252 | -0.539 | 0.274 | -0.193 |
| log.spec.orig.main | 0.360 | 0.363 | 0.229 | -0.563 | 0.199 | -0.154 |
| fro.over.spec | -0.290 | -0.298 | 0.275 | -0.586 | 0.315 | -0.354 |
| **Frobenius-norm measures** | | | | | | |
| fro.dist | 0.165 | 0.139 | -0.114 | -0.094 | -0.122 | 0.052 |
| log.prod.of.fro | -0.197 | -0.299 | 0.320 | -0.469 | 0.390 | -0.364 |
| log.prod.of.fro.over.margin | -0.216 | -0.295 | 0.320 | -0.469 | 0.357 | -0.349 |
| log.sum.of.fro | -0.155 | -0.251 | 0.275 | -0.516 | 0.325 | -0.338 |
| log.sum.of.fro.over.margin | -0.193 | -0.272 | 0.275 | -0.516 | 0.092 | -0.116 |
| log.dist.spec.init | 0.100 | 0.069 | -0.503 | 0.141 | -0.239 | 0.158 |
| param.norm | -0.211 | -0.288 | 0.366 | -0.422 | -0.214 | 0.120 |
| **Path-based measures** | | | | | | |
| path.norm | 0.234 | 0.288 | -0.023 | 0.164 | -0.357 | 0.339 |
| param.norm.over.margin | 0.281 | 0.331 | -0.092 | 0.141 | -0.436 | 0.401 |
| **Flatness-based measures** | | | | | | |
| pacbayes.flatness | 0.327 | 0.373 | -0.211 | 0.144 | 0.417 | -0.306 |
| pacbayes.init | 0.248 | 0.277 | -0.160 | -0.070 | 0.316 | -0.312 |
| pacbayes.orig | -0.053 | -0.048 | 0.206 | -0.164 | 0.296 | -0.306 |
| pacbayes.mag.flatness | 0.491 | 0.533 | -0.372 | 0.381 | -0.342 | 0.211 |
| pacbayes.mag.init | -0.276 | -0.297 | 0.320 | -0.469 | 0.389 | -0.387 |
| pacbayes.mag.orig | -0.285 | -0.299 | 0.320 | -0.469 | 0.387 | -0.385 |
| **Optimization measure** | | | | | | |
| #steps.%error | 0.323 | 0.330 | -0.275 | 0.435 | -0.462 | 0.378 |

# 4 Discussion

In this work, we attempt to acquire insights on whether complexity measures can predict the impact of deep network architectures on their generalization performance on medical images. Toward this goal, we performed experiments over a space of hypotheses. In conclusion, the measure that correlated the most consistently across the experiments in this study is the magnitude-aware PAC-Bayes flatness measure (with perturbation loss set to 0.1). Other measures that exhibited positive correlation with generalization are the two path-based measures, inverse margin, and optimization-based number of train steps. For the predictors trained to zero empirical risk, almost all spectral norm-based measures showed strong correlation. Some spectral norm measures resulted in positive correlation over all experiments (e.g., log.sum.of.spec.over.margin), while most Frobenius norm-based measures were negatively correlated with generalization (except for low correlation of log.dist.spec.init).

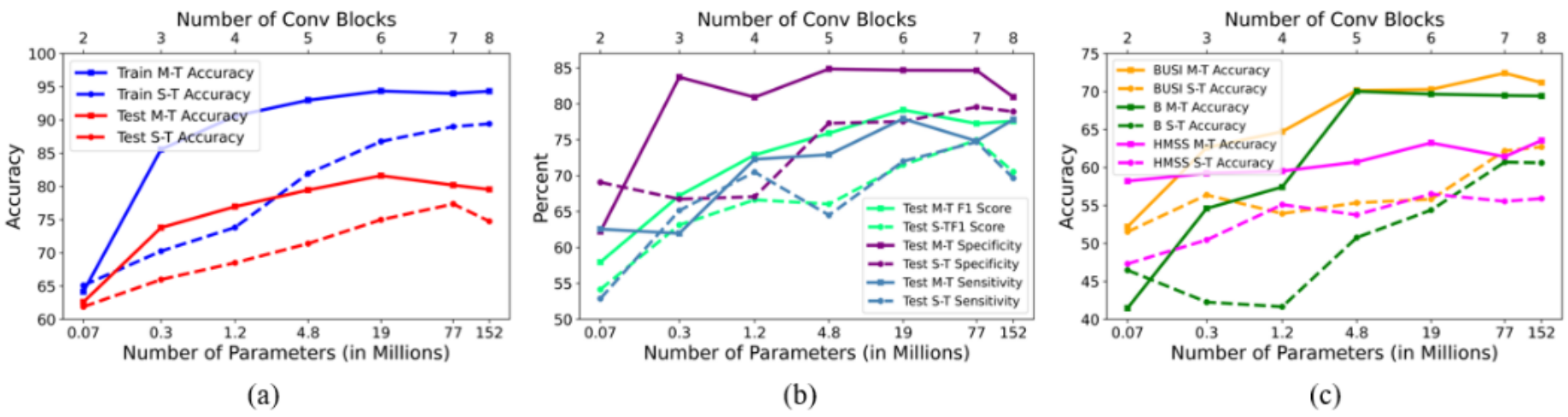

**Fig. 3.** Comparison of multi-task (M-T, solid lines) and single-task (S-T, dashed lines) predictors: (a) Train and test accuracy on BUSIS dataset; (b) F1 score, specificity, and sensitivity on BUSIS test dataset; (c) Accuracy on BUSI, B, and HMSS datasets.

Differently from the similar empirical studies in [16] and [18], we evaluated VGG-like networks (rather than NiN architectures) on small medical datasets with larger-resolution images. We also studied predictors that (similar to current practice) are not trained to full convergence and apply regularization, e.g., via early stopping, batch-norm, dropout layers: the results indicated that the evaluation of complexity measures using the test risk is more consistent with related studies. Also, unlike [16] and [18], we report that most spectral-norm measures exhibit positive correlation with generalization. In addition, we investigated the effectiveness of multi-task learning with BUS images, and report improved generalization on both i.i.d. and o.o.d. images.

In future work, we plan to evaluate the influence of other hyperparameters, optimization algorithms, and network architectures. The findings of ours and related studies can be used for model selection, architecture search, and improved interpretability and robustness of deep learning-based methods for medical image analysis.

**Acknowledgment.** Supported by the Center for Modeling Complex Interactions at the University of Idaho through National Institutes of Health Award P20GM104420.

# Appendix

## A.1 Network Architectures

As explained in the Methods section, the models in the experiments use a series of Conv Blocks, followed by a Fully-Connected Block (see Fig. 1). The layers in the Conv Blocks are described in Table A1. Each Conv Block consists of two convolutional layers, and/or a max pooling layer. For input images resized to 256×256 pixels (having a single channel), the size of the output feature maps from each block is shown in the right column in table A1. For the networks with more than 5 Conv Blocks, a max pooling layer was not applied, since we noticed decreased performance when the size of the extracted feature maps was reduced to less than 8×8 pixels. In addition, unlike Conv Blocks 1 to 7, the number of convolutional layers in Conv Block 8 was not doubled, because the size of the model with 4,096 filters crashed the memory when calculating the norms for the complexity measures.

The upconvolutional (UpConv) Blocks used in the decoder branch of the joint classification and segmentation models (Fig. 1) have the same composition as the corresponding Conv Blocks in the encoder branch of the classification models.

**Table A1.** Description of the layers in the Conv Blocks used in the networks. The size of input images is 256×256 pixels.

| Block | Convolutional Layer | Convolutional Layer | Max Pooling Layer | Size of Output Feature Maps |
|---|---|---|---|---|
| Conv Block 1 | 32 conv 3×3 filters | 32 conv 3×3 filters | 2×2 | 128×128×32 |
| Conv Block 2 | 64 conv 3×3 filters | 64 conv 3×3 filters | 2×2 | 64×64×64 |
| Conv Block 3 | 128 conv 3×3 filters | 128 conv 3×3 filters | 2×2 | 32×32×128 |
| Conv Block 4 | 256 conv 3×3 filters | 256 conv 3×3 filters | 2×2 | 16×16×256 |
| Conv Block 5 | 512 conv 3×3 filters | 512 conv 3×3 filters | 2×2 | 8×8×512 |
| Conv Block 6 | 1,024 conv 3×3 filters | 1,024 conv 3×3 filters | | 8×8×1,024 |
| Conv Block 7 | 2,048 conv 3×3 filters | 2,048 conv 3×3 filters | | 8×8×2,048 |
| Conv Block 8 | 2,048 conv 3×3 filters | 2,048 conv 3×3 filters | | 8×8×2,048 |

**Table A2.** Network architecture details: number of blocks, total number of trainable parameters, and order of the convolutional blocks and fully-connected layers.

| Number of Blocks | Number of Parameters | Conv Blocks | Fully-connected Layers |
|---|---|---|---|
| 2 Blocks | 67,968 | Conv Block 1 to 2 | 32–16–2 |
| 3 Blocks | 296,864 | Conv Block 1 to 3 | 64–32–2 |
| 4 Blocks | 1,212,384 | Conv Block 1 to 4 | 128–64–2 |
| 5 Blocks | 4,857,824 | Conv Block 1 to 5 | 256–64–2 |
| 6 Blocks | 19,423,200 | Conv Block 1 to 6 | 512–64–2 |
| 7 Blocks | 76,570,592 | Conv Block 1 to 7 | 512–64–2 |
| 8 Blocks | 152,068,064 | Conv Block 1 to 8 | 512–64–2 |

Additional information about the network architectures is provided in Table A2. I.e., the overall number of trainable parameters for each architecture is presented, as well as the used Conv Blocks and Fully-Connected (FC) Blocks. The FC Blocks consist of three fully-connected layers, with the number of computational units in each architecture shown in the right column in Table A2. The number of computational units in the FC Block is varied for each architecture.

The models were implemented in PyTorch using the Google Colaboratory Pro cloud computing services, which typically provide access to NVIDIA V100 GPUs.

### A.2 Complexity Measures

A brief description of the used complexity measures follows. A full description can be found in [16], and partially in [18].

**VC dimension-based measure.** This measure corresponds to the number of parameters of a deep convolutional network. For a network with a number of layers $d$, layer index $i$, and a number of convolutional filters $c_i$ at layer $i$ with a kernel size $k_i \times k_i$, it is:

$$\mu_{\text{num.params}} = \sum_i k_i^2 c_{i-1}(c_i + 1) \quad \text{(A1)}$$

**Output-based measure.** It is a data-dependent measures, based on the squared inverse of the margin γ. The margin is obtained as the 10th percentile of the distribution of the differences in the logit values between the binary outputs over the training dataset. Many of the remaining complexity measures employ the margin $\gamma$ as a normalization constant.

$$\mu_{\text{inverse.margin}} = 1/\gamma^2 \quad \text{(A2)}$$

**Spectral norm-based measures.** We used 7 related measures, where the weight tensors of the $i^{\text{th}}$ layer of a trained model are denoted $\mathbf{W}_i$, and the initial weight tensors of the $i^{\text{th}}$ layer before the training are $\mathbf{W}_i^0$. Similarly, the notation $\|\cdot\|_2$ is used for the spectral norm of a matrix, and $\|\cdot\|_{\mathrm{F}}$ is used for the Frobenius norm. The complexity measures are given by:

$$\mu_{\text{log.prod.of.spec}} = \log\left(\prod_i \|\mathbf{W}_i\|_2^2\right) \quad \text{(A3)}$$

$$\mu_{\text{log.prod.of.spec.over.margin}} = \log\left(\prod_i \|\mathbf{W}_i\|_2^2/\gamma^2\right) \quad \text{(A4)}$$

$$\mu_{\text{log.sum.of.spec}} = \log\left(d\left(\prod_i \|\mathbf{W}_i\|_2^2\right)^{1/d}\right) \quad \text{(A5)}$$

$$\mu_{\text{log.sum.of.spec.over.margin}} = \log\left(d\left(\prod_i \|\mathbf{W}_i\|_2^2/\gamma^2\right)^{1/d}\right) \quad \text{(A6)}$$

$$\mu_{\text{log.spec.init.main}} = \log\left(\prod_i \|\mathbf{W}_i\|_2^2 \sum_j \left(\left\|\mathbf{W}_j - \mathbf{W}_j^0\right\|_F^2\right)/\left\|\mathbf{W}_j\right\|_2^2/\gamma^2\right) \quad \text{(A7)}$$

$$\mu_{\text{log.spec.orig.main}} = \log\left(\prod_i \|\mathbf{W}_i\|_2^2 \sum_j \left(\left\|\mathbf{W}_j\right\|_F^2\right)/\left\|\mathbf{W}_j\right\|_2^2/\gamma^2\right) \quad \text{(A8)}$$

$$\mu_{\text{fro.over.spec}} = \sum_i \left(\|\mathbf{W}_i\|_F^2\right)/\|\mathbf{W}_i\|_2^2 \quad \text{(A9)}$$

**Frobenius norm-based measures.** Analogously, we evaluated the following 7 complexity measures that are established using Frobenius norms of the model weights.

$$\mu_{\text{fro.dist}} = \sum_i \|\mathbf{W}_i - \mathbf{W}_i^0\|_F^2 \tag{A10}$$

$$\mu_{\text{log.prod.of.fro}} = \log\left(\prod_i \|\mathbf{W}_i\|_F^2\right) \tag{A11}$$

$$\mu_{\text{log.prod.of.fro.over.margin}} = \log\left(\prod_i \|\mathbf{W}_i\|_F^2 / \gamma^2\right) \tag{A12}$$

$$\mu_{\text{log.sum.of.fro}} = \log\left(d\left(\prod_i \|\mathbf{W}_i\|_F^2\right)^{1/d}\right) \tag{A13}$$

$$\mu_{\text{log.sum.of.fro.over.margin}} = \log\left(d\left(\prod_i \|\mathbf{W}_i\|_F^2 / \gamma^2\right)^{1/d}\right) \tag{A14}$$

$$\mu_{\text{dist.spec.init}} = \sum_i \|\mathbf{W}_i - \mathbf{W}_i^0\|_2^2 \tag{A15}$$

$$\mu_{\text{param.norm}} = \sum_i \|\mathbf{W}_i\|_F^2 \tag{A16}$$

**Path-based measures.** These two measures are based on the values of the logits outputs when using squared weights $\mathbf{w}^2$ where $\mathbf{w} = \text{vec}(\mathbf{W}_1, \ldots, \mathbf{W}_d)$, and the inputs are vectors of ones, i.e., $h_{\mathbf{w}^2}(\mathbf{1})$.

$$\mu_{\text{path.norm}} = \sum_i h_{\mathbf{w}^2}(\mathbf{1})\,[i] \tag{A17}$$

$$\mu_{\text{path.norm.over.margin}} = \left(\sum_i h_{\mathbf{w}^2}(\mathbf{1})[i]\right) / \gamma^2 \tag{A18}$$

**Flatness-based measures.** The complexity measures in this group are derived based on the PAC-Bayes theory. By adding Gaussian perturbations $\mathbf{u} \sim \mathcal{N}(0, \sigma^2 I)$ to the weights of a trained model, the measures are calculated based on the variance $\sigma^2$ that is chosen to be the largest number that causes the expected empirical risk on the training data to reduce by at least 10%, i.e., $E_{\mathbf{u} \sim \mathcal{N}(0,\sigma^2 I)}\left[\hat{L}(h_{\mathbf{w}+\mathbf{u}})\right] \leq 0.1$. The magnitude-aware measures apply perturbations $\mathbf{u} \sim \mathcal{N}(0, \sigma'^2 |w_i|^2 + \epsilon^2)$ that are adjusted to the magnitude of the weights, where the variance $\sigma'^2$ is chosen to be the largest number such that $E_{\mathbf{u} \sim \mathcal{N}(0,\sigma'^2 |\mathrm{w}_i|^2 + \epsilon^2)}\left[\hat{L}(h_{\mathbf{w}+\mathbf{u}})\right] \leq 0.1$. The value $\epsilon$ was set to $10^{-3}$. In the following equations $\omega$ denotes the number of weights.

$$\mu_{\text{pacbayes.flatness}} = 1/\sigma^2 \tag{A19}$$

$$\mu_{\text{pacbayes.init}} = \|\mathbf{w} - \mathbf{w}^0\|_2^2 / 4\sigma^2 + \log((m+2)/\delta) \tag{A20}$$

$$\mu_{\text{pacbayes.orig}} = \|\mathbf{w}\|_2^2 / 4\sigma^2 + \log((m+2)/\delta) \tag{A21}$$

$$\mu_{\text{pacbayes.mag.flatness}} = 1/\sigma'^2 \tag{A22}$$

$$\mu_{\text{pacbayes.mag.init}} = (1/4) \sum_i \log\Big((\epsilon^2 + (\sigma'^2 + 1)\, \|\mathbf{w} - \mathbf{w}^0\|_2^2 / \omega) / \ldots$$

$$\left(\epsilon^2 + \sigma'^2 \left|\omega_i - \omega_i^0\right|^2\right)\Big) + \log((m+2)/\delta) \tag{A23}$$

$$\mu_{\text{pacbayes.mag.orig}} = (1/4)\sum_i \log\Big((\epsilon^2 + (\sigma'^2+1)\,\|\mathbf{w}\|_2^2/\omega)/\ldots$$

$$\Big(\epsilon^2 + \sigma'^2\big|\omega_i - \omega_i^0\big|^2\Big)\Big) + \log((m+2)/\delta) \quad \text{(A24)}$$

**Optimization-based measure.** We used the number of steps for achieving a training classification error of either 0.01 (for experiment $E_1$) or 0.1 (for experiments $E_2$ and $E_3$).

$$\mu_{\#\text{steps.\%error}} = \text{number of steps to 0.01 or 0.1\% error} \quad \text{(A25)}$$

### A.3 Experimental Results

The test and train accuracies for single-task (experiment $E_2$) and multi-task (experiment $E_3$) learning approaches are presented in Table A3. These values correspond to the points in the graphs in Fig. 3a (as well as, are related to the graphs showing the errors in Fig. 2bc). The values of the F1 score, sensitivity, and specificity for the multi-task approach are also shown in Table A3 (corresponding to Fig. 3b).

Similarly, the classification accuracies for the single-task and multi-task models trained on the BUSIS dataset and evaluated on the BUSI, B, and HMSS datasets are presented in Table A4 (the values correspond to the graph lines shown in Fig. 3c).

**Table A3.** Test and train accuracies (mean and standard deviation) for single-task (S-T) and multi-task (M-T) predictors, F1 scores, sensitivity, and specificity using the BUSIS dataset.

| Blocks | Test S-T | Train S-T | Test M-T | Train M-T | F1 Score M-T | Sensitivity M-T | Specificity M-T |
|---|---|---|---|---|---|---|---|
| 2 | 61.89 (6.8) | 65.10 (4.6) | 62.60 (5.2) | 64.17 (5.6) | 57.96 (13.0) | 62.56 (24.4) | 62.21 (22.0) |
| 3 | 65.98 (5.1) | 70.26 (5.2) | 73.80 (4.9) | 85.56 (5.3) | 67.22 (9.6) | 61.96 (16.2) | 83.72 (9.8) |
| 4 | 68.51 (5.1) | 73.83 (6.0) | 76.96 (5.2) | 90.61 (5.6) | 72.88 (10.0) | 72.28 (17.2) | 80.92 (11.4) |
| 5 | 71.40 (4.4) | 81.94 (6.4) | 79.46 (5.2) | 92.98 (4.5) | 75.90 (7.6) | 72.93 (12.9) | 84.86 (11.9) |
| 6 | 74.99 (3.5) | 86.77 (4.3) | **81.62** (4.6) | **94.37** (2.6) | **79.16** (6.6) | **77.93** (10.0) | **84.69** (8.0) |
| 7 | **77.36** (5.3) | 89.00 (3.2) | 80.20 (3.9) | 94.00 (1.6) | 77.26 (5.2) | 74.81 (8.6) | 84.65 (5.6) |
| 8 | 74.76 (4.5) | **89.43** (3.8) | 79.55 (4.0) | 94.34 (1.1) | 77.60 (4.1) | 77.80 (5.6) | 80.95 (6.7) |

**Table A4.** Accuracies (mean and standard deviation) for single-task (S-T) and multi-task (M-T) predictors using the BUSI, B, and HMSS datasets.

| Blocks | BUSI S-T | BUSI MT | B S-T | B M-T | HMSS S-T | HMSS M-T |
|---|---|---|---|---|---|---|
| 2 | 51.53 (8.7) | 52.19 (11.6) | 46.47 (6.5) | 41.47 (9.3) | 47.32 (2.5) | 58.21 (3.6) |
| 3 | 56.38 (9.5) | 62.62 (6.2) | 42.27 (5.9) | 54.60 (8.1) | 50.46 (3.7) | 59.24 (2.4) |
| 4 | 53.93 (9.8) | 64.72 (7.0) | 41.66 (7.9) | 57.42 (7.5) | 55.11 (2.6) | 59.52 (2.6) |
| 5 | 55.34 (12.9) | 70.15 (8.6) | 50.77 (13.1) | **70.03** (7.5) | 53.76 (4.1) | 60.72 (2.6) |
| 6 | 55.82 (12.6) | 70.28 (7.8) | 54.39 (11.1) | 69.66 (5.6) | **56.51** (4.5) | 63.24 (3.1) |
| 7 | 62.19 (7.9) | **72.42** (4.0) | **60.71** (10.2) | 69.48 (5.3) | 55.54 (4.1) | 61.43 (5.4) |
| 8 | **62.70** (9.6) | 71.18 (5.0) | 60.61 (7.8) | 69.42 (4.3) | 55.91 (4.0) | **63.56** (2.4) |

## A.4 Robust Kendall's Sign-ranking Coefficients

The results of the correlation analysis between the complexity measures and generalization performance for the predictors in experiment $E_1$ based on distributional robustness [18] are presented in Fig. 4A. The bars in the figure depict the mean, maximum, and 90$^{th}$ percentile values of the weighted sign-errors for each complexity measure. Note that conventional Kendall's coefficients employ sign-ranking of the concordant and discordant pairs of observations, are in the [-1, +1] range, and higher coefficients indicate high correlation. Differently, the robust Kendall's coefficients as proposed in [18] are calculated based on the sign-error of discordant pairs of observations, are scaled and have values in the [0, +1] range, and lower coefficients indicate high correlations. E.g., the magnitude-aware PAC-Bayes flatness measures achieved the best overall correlation with generalization, which is in agreement with the results for the Kendall's and granulated Kendall's coefficients. The figure provides more details regarding the correlations: e.g., one can notice that many complexity measures have a maximum value of 1, meaning that at least for one pair of predictors groups $H_1$ and $H_2$, the measures failed to capture the generalization behavior.

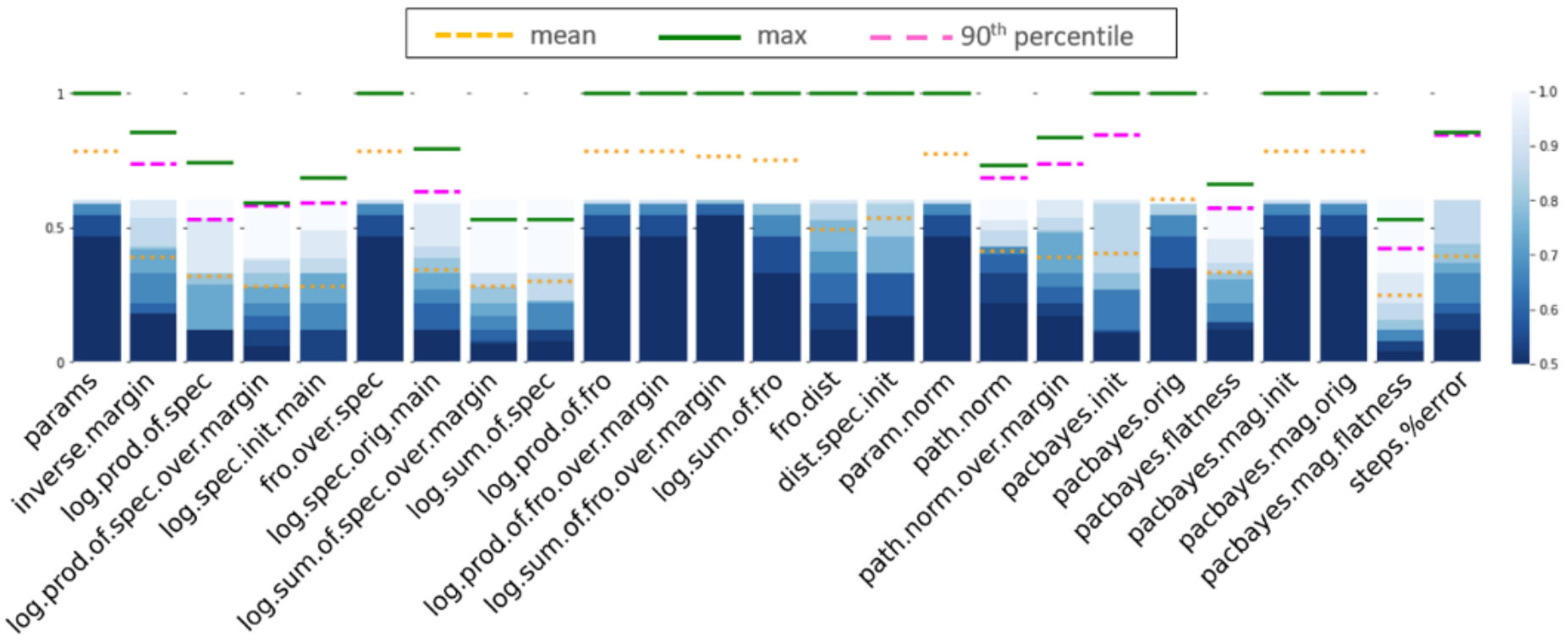

**Fig. 4A.** Robust Kendall's coefficient for the experiment $E_1$ (predictors trained to full convergence). The graphs show the mean, maximum, and 90$^{th}$ percentile values of the weighted sign-errors for all complexity measures (lower is better). The bar colors depict the cumulative distribution function of the sign-errors: white indicates robustness, blue is reduced robustness.

The colors in the bars depict the cumulative distribution function (CDF) for the complexity measures, where white color represents improved robustness, and blue color represents reduced robustness. Conclusively, Frobenius norm-based measures are the least robust, some spectral norm-based measures demonstrate robustness, and the flatness- and path-based measures are the most robust.